\documentclass{article}

\usepackage{acra}
\usepackage{graphicx}
\usepackage{tabularx}
\usepackage{longtable}
\usepackage{nicematrix}
\usepackage{tikz}
\usepackage{subcaption}
\usepackage{float}
\usepackage{hyperref}
\usepackage{caption}
\usepackage{bbding}
\usepackage{pifont}
\usepackage{wasysym}
\usepackage{amssymb}
\usepackage{amsmath,amsthm}
\usepackage{multicol}
\usepackage{cuted}
\usepackage{multirow}
\usepackage{changepage}
\usepackage{balance} 
\usepackage{flushend}
\title{Australian Supermarket Object Set (ASOS): A Benchmark Dataset of Physical Objects and 3D Models for Robotics and Computer Vision}

\author{
  \begin{tabular}{ccc}
    Akansel Cosgun & Lachlan Chumbley & Benjamin J. Meyer \\
    Deakin University, Australia & Monash University, Australia & Coles Group, Australia
  \end{tabular}
}

\begin{document}

\maketitle

\begin{strip}
\vspace{-1.5cm}  
\centering
\includegraphics[width=\textwidth]{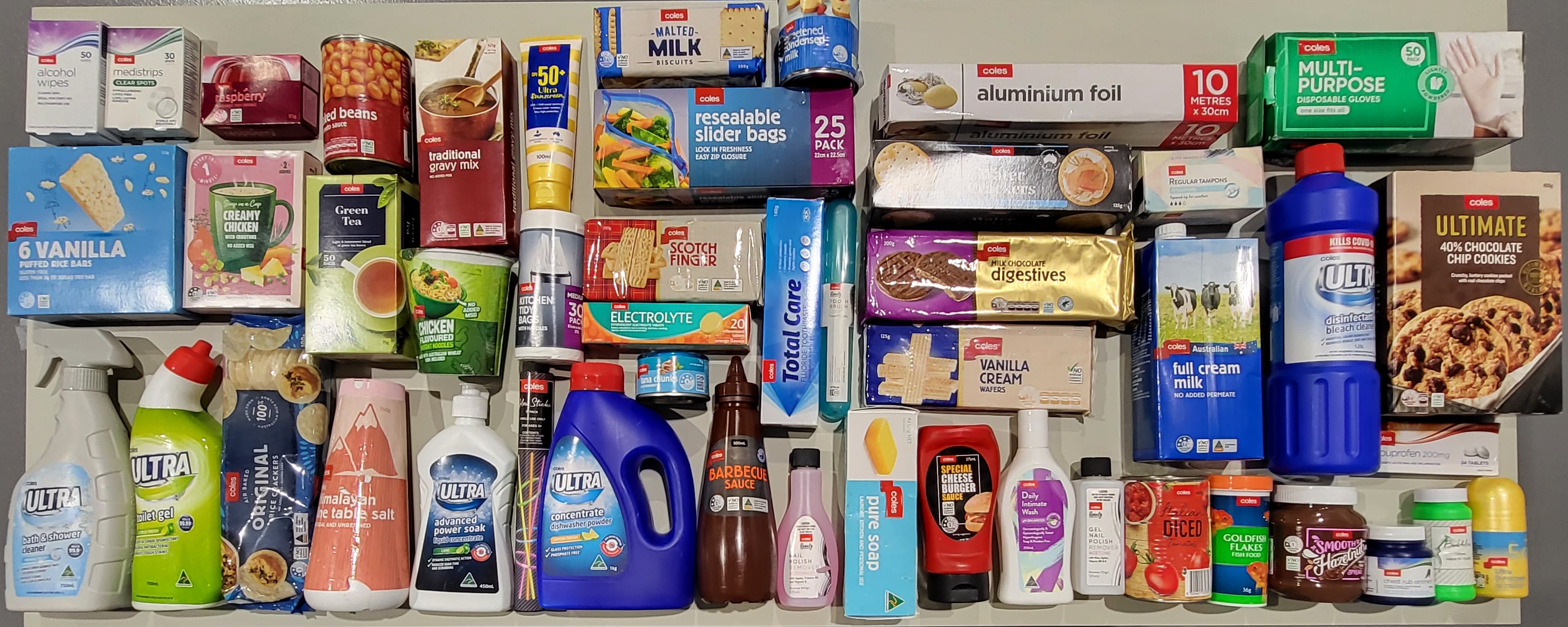}
\captionof{figure}{Australian Supermarket Object Set (ASOS): 50 objects with 3D textured meshes.}
\label{fig:intro}
\end{strip}


\begin{abstract}
This paper introduces the Australian Supermarket Object Set (ASOS), a comprehensive dataset comprising 50 readily available supermarket items with high-quality 3D textured meshes designed for benchmarking in robotics and computer vision applications. Unlike existing datasets that rely on synthetic models or specialized objects with limited accessibility, ASOS provides a cost-effective collection of common household items that can be sourced from a major Australian supermarket chains. The dataset spans 10 distinct categories with diverse shapes, sizes, and weights. 3D meshes are acquired by a structure-from-motion techniques with high-resolution imaging to generate watertight meshes. The dataset's emphasis on accessibility and real-world applicability makes it valuable for benchmarking object detection, pose estimation, and robotics applications.
\end{abstract}

\section{Introduction}

Standardized object datasets like the YCB Object and Model Set \cite{calli2015ycb} are crucial for benchmarking in computer vision and robotics, providing consistent evaluation metrics and reproducible results. While the YCB set's 77 objects across food, tools, and task items have proven valuable for tasks from recognition to pose estimation, international researchers often face accessibility challenges when attempting to acquire the complete set. Additionally, there remains a notable gap in standardized datasets for supermarket objects with varying deformability and mass distributions.

In response to these challenges, we introduce the Australian Supermarket Object Set (ASOS)\footnote{\href{https://lachlanchumbley.github.io/ColesObjectSet/}{https://lachlanchumbley.github.io/ColesObjectSet/}.} (Figure \ref{fig:intro}), designed for accessibility and relevance to real-world scenarios. This dataset contains 50 objects commonly found in Australian supermarkets, along with their 3D textured meshes. These objects were chosen based on their affordability, availability, and representativeness of typical household items, ensuring their usability for both real-world and simulated environments. Unlike many existing datasets that rely on synthetic models or lack real-world counterparts, the supermarket object set provides tangible objects that can be easily sourced locally, overcoming barriers of cost and availability while addressing practical issues like deformability and mass distribution that are difficult to replicate in simulations.

The availability of this object set addresses a gap in the existing literature: while datasets like LINEMOD \cite{Hinterstoier2012ModelBT} and BigBIRD \cite{6906903} focus on pose estimation and RGB-D tasks, their reliance on online meshes or specialized items limits their real-world applicability. Similarly, while the YCB and ACRV datasets \cite{7989545} enable real-world object benchmarking, their emphasis on general household items leaves room for a domain-specific dataset that captures objects encountered in daily shopping contexts. The Australian Supermarket Object Set bridges this gap by offering a cost-effective, practical, and diverse collection of items that can be leveraged to evaluate robotics and computer vision algorithms in real-world and simulated environments.

This paper outlines the design principles, data collection methodology, and metadata associated with the supermarket object set. By providing an open-access dataset and addressing challenges like sim-to-real discrepancies, we aim to advance reproducibility and innovation in robotics and vision research.

\begin{table*}

\setlength{\tabcolsep}{2pt}
\centering


\newcolumntype{C}[1]{>{\PreserveBackslash\centering}p{#1}}
\centering
\begin{tabular}{|p{0.33\textwidth}|p{0.34\textwidth}|p{0.13\textwidth}|p{0.042\textwidth}|p{0.11\textwidth}|}
      \hline \textbf{Dataset Name} & \textbf{Type} & \textbf{Theme} & \textbf{\# objs} & \raggedright{\textbf{Real objects?}}\arraybackslash \\ \hline
      PSB \cite{10.5555/998687.1007045}  & Meshes & General & 1814 & \\ \hline
      3DNet \cite{6225116} & Meshes & General & 3433 &  \\ \hline
      KIT database \cite{doi:10.1177/0278364912445831}  & Textured meshes, stereo RGB images & Household & 145 &  \\ \hline
      LINEMOD \cite{Hinterstoier2012ModelBT}  & Textured meshes, RGB images with poses & General & 15 &  \\ \hline
      BigBIRD \cite{6906903}  & Textured meshes, RGB-D images & Household & 125 &  \\ \hline
      ModelNet \cite{Zhirong15CVPR}  & Meshes & General & 151k &  \\ \hline
      Rutgers APC \cite{Rennie2015ADF}  & Textured meshes, RGB-D images with poses & Household & 25 &  \\ \hline
      ShapeNetCore \cite{shapenet2015}  & Meshes with WordNet annotations & General & 51k &  \\ \hline
      YCB \cite{calli2015ycb}  & Textured meshes, RGB-D images, shopping list & Daily life & 77 & \checkmark \\ \hline
      ACRV \cite{7989545}  & Textured meshes, shopping list & Household & 42 & \checkmark \\ \hline
      MVTec ITODD \cite{8265467}  & Meshes, RGB-D images with poses & Industrial & 28 &  \\ \hline
      T-LESS \cite{hodan2017tless}  & Textured meshes, RGB-D images with poses & Industrial & 30 &  \\ \hline
      RBO \cite{doi:10.1177/0278364919844314}  & Articulated meshes, RGB-D images & Articulation & 14 &  \\ \hline
      TUD-L \& TYO-L \cite{hodan2018bop}  & Textured meshes, RGB-D images with poses & Varied lighting & 24 &  \\ \hline
      ContactDB \cite{Brahmbhatt2019ContactDBAA}  & Meshes with contact maps, RGB-D \& thermal images & Contact & 3750 & 3D printable \\ \hline
      HomebrewedDB \cite{9021987}  & Textured meshes, RGB-D images with poses & Household, industry, toy & 33 &  \\ \hline
      EGAD \cite{morrison2020egad}  & Meshes, 3D printing instructions & Generated & 2282 & 3D printable \\ \hline
      Household Cloth Object Set \cite{9732698}  & Meshes, microscopic images, object details & Cloth & 27 & \checkmark \\ \hline
      ABO \cite{collins2022abo}  & Textured meshes, RGB images, physically-based renders & Amazon.com household & 7953 &  \\ \hline
      AKB-48 \cite{Liu2022AKB48AR}  & Articulated meshes & Articulation & 2037 &  \\ \hline
      GSO \cite{10.1109/ICRA46639.2022.9811809}  & High-quality textured meshes with metadata  & Household & 1030 &  \\ \hline
      HOPE \cite{tyree2022hope}  & Textured meshes, shopping list & Toy Grocery & 28 & $\checkmark$ \\ \hline
      MP6D \cite{9722997}  & Meshes, RGB-D images & Industrial & 20 &  \\ \hline
      ObjectFolder \cite{gao2022ObjectFolderV2}\cite{gao2021ObjectFolder}  & Neural representation for visual, impact sounds and tactile data & Multisensory & 1000 &  \\ \hline
      PCPD \cite{9959243}  & RGB-D images & Power grid & 10 &  \\ \hline
      TransCG \cite{fang2022transcg}  & Meshes, RGB-D images & Transparent & 51 &  \\ \hline
      \textbf{Ours} & \textbf{Textured meshes, shopping list} & \textbf{Supermarket} & \textbf{50} & \checkmark \\ \hline
\end{tabular}
\caption{Object datasets present in the literature. (Real objects available) We consider objects to be `3D printable' if the dataset presented the 3d printed models as the object dataset, compared to the dataset containing 3D printable objects that would produce objects with different properties than those in the dataset. Similarly, we in general consider real objects to be available if they can be obtained in their original form without much effort, and for a reasonable monetary cost, based solely on the information provided by the authors of the dataset.}
\label{tab:related_objects}
\end{table*}

\section{Related Works}
Object sets have been widely used as benchmarking tools for various tasks. One of the oldest and still commonly used object sets is the Princeton Shape Benchmark \cite{10.5555/998687.1007045}, which focuses on a range of geometries with semantic labels indicating their purpose. However, this object set only provides untextured meshes and is primarily used for shape-based tasks such as geometric matching. Other object sets like 3DNet \cite{6225116}, ModelNet \cite{Zhirong15CVPR}, and ShapeNet \cite{shapenet2015} emphasize geometrical properties and lack texture. While these sets contain semantically labeled CAD models, a major limitation of such object sets is that there are no guarantees about how models in these object sets correspond to real-world objects. These object sets are usually curated by humans to eliminate unrealistic geometries and ensure correct semantic labels, but many of the models in these databases were synthetically constructed and so may not map to real world objects. Additionally, the absence of texture hinders the integration of visual cues for tasks like segmentation and pose estimation.

There have been many other object sets proposed for use in benchmarking object detection and pose estimation, among other tasks, that consist of textured object models that have direct real-world analogs. Early object sets introduced with these features include LINEMOD\cite{Hinterstoier2012ModelBT}, the KIT database \cite{doi:10.1177/0278364912445831} and BigBIRD \cite{6906903}, with more recently introduced datasets improving on these early benchmarks, such as GSO \cite{10.1109/ICRA46639.2022.9811809} consisting of much higher quality scans and ABO \cite{collins2022abo} containing a wide range of realistic textured meshes. There have also been many object sets proposed that are more focused on a specific task, which include extra data that is directly relevant to the task. Object sets that were designed primarily for pose estimation such as the Rutgers APC RGB-D dataset \cite{Rennie2015ADF}, TUD-L, TYO-L \cite{hodan2018bop} and HomebrewedDB \cite{9021987}, contain RGB-D images that have been annotated with object poses in addition to textured object meshes. 

All object sets that have been mentioned so far consist of meshes only accessible online. While many of these object sets may have been constructed by scanning real-world objects, there is no easy way to access the original objects used to construct the object sets to perform real-world tests. This is especially an issue for robotics tasks, where direct interaction with the object is commonplace, and there can often be a large sim-to-real gap due to many properties of real-world objects not being accounted for by the simulated meshes, such as surface roughness and deformability. To rectify this issue, there have been object sets introduced that aim to provide access to the real world objects that correspond to the meshed in the datasets for use in robotic benchmarking of tasks like grasping. Two primary examples of such object sets are the YCB Object and Model Set \cite{calli2015ycb} and ACRV \cite{7989545}, which consist of household objects and come with shopping lists to allow researchers to easily purchase the real-world objects.

\section{Supermarket Object Set}
\label{sec:object_set}

The Supermarket Object Set is a collection of 50 household items that can be easily obtained from Coles, a major Australian supermarket chain. This object set data is accessible online\footnote{\href{https://lachlanchumbley.github.io/ColesObjectSet/}{https://lachlanchumbley.github.io/ColesObjectSet/}.}. Each object in the set is accompanied by a high-quality 3D water-tight mesh, as well as detailed information about its mass and dimensions. This object set is specifically designed to facilitate benchmarking of robotic manipulation and computer vision tasks, providing researchers with accessible and common objects for evaluation. The inclusion of high-quality meshes in the Supermarket Object Set enables accurate simulation of real-world objects, allowing for benchmarking of manipulation and vision techniques. However, it is important to note that certain properties of real objects, such as flexibility, deformability, and durability, are challenging to simulate effectively. Additionally, properties like mass distribution are difficult to measure accurately for real objects. The accessibility of the real-world object set provides an opportunity to compare algorithms in these hard-to-simulate circumstances. 

The rest of this section presents how the objects were chosen (Section \ref{sec:object_choices}), the data collection method (Section \ref{sec:data_collection}) and the metadata (Section \ref{sec:metadata}).

\subsection{Object Choices}
\label{sec:object_choices}

\begin{figure*}
    \scriptsize
    \def\arraystretch{-3} 
    \captionsetup{justification=centering}
    \begin{NiceTabularX}{\textwidth}{X[1,c]X[1,c]X[1,c]X[1,c]X[1,c]X[1,c]}
        \includegraphics[width=0.6\textwidth,trim={4cm 2cm 4cm 2cm},clip]{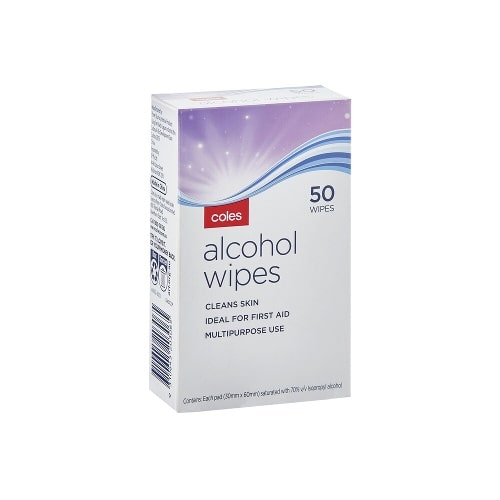}\break Alcohol Wipes&
        \includegraphics[width=0.6\textwidth,trim={2cm 2cm 2cm 2cm},clip]{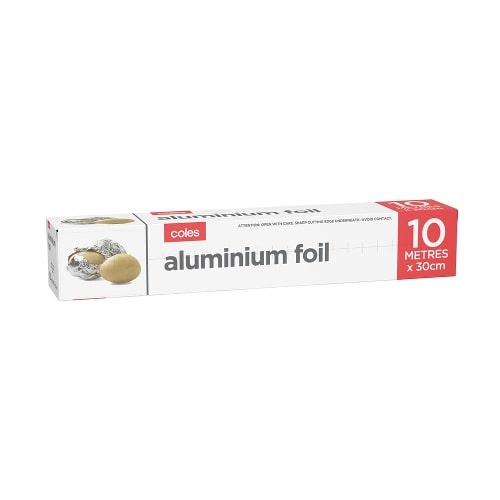}\break{Aluminium Foil}&
        \includegraphics[width=0.6\textwidth,trim={4cm 2cm 4cm 2cm},clip]{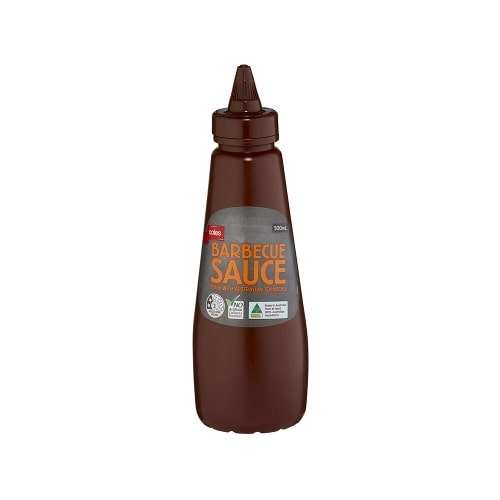}\break{BBQ Sauce}&
        \includegraphics[width=0.6\textwidth,trim={4cm 2cm 4cm 2cm},clip]{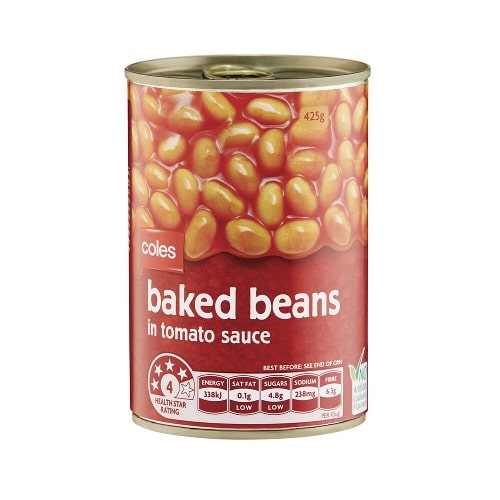}\break{Baked Beans}&
        \includegraphics[width=0.6\textwidth,trim={4cm 2cm 4cm 2cm},clip]{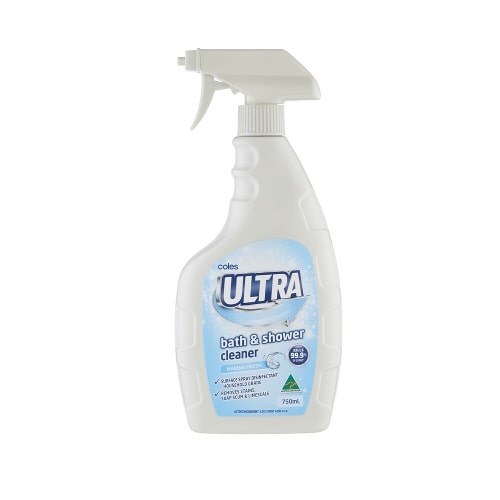}\break{Bathroom Cleaner}&
        \includegraphics[width=0.6\textwidth,trim={4cm 1cm 4cm 2cm},clip]{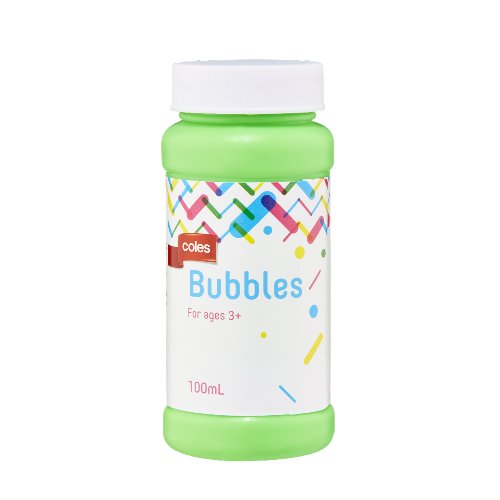}\break{Bubbles}\\
        \includegraphics[width=0.6\textwidth,trim={4cm 2cm 4cm 2cm},clip]{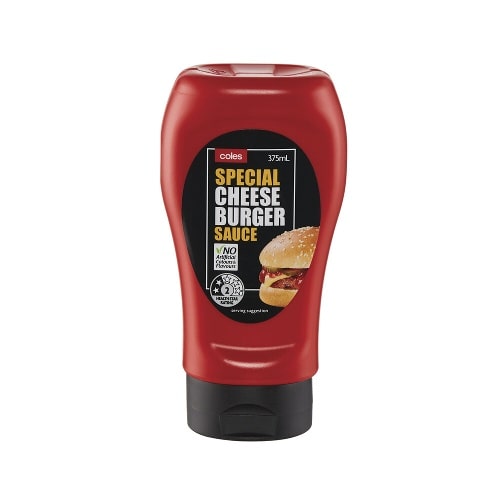}\break{Burger Sauce}&
        \includegraphics[width=0.6\textwidth,trim={4cm 2cm 3cm 2cm},clip]{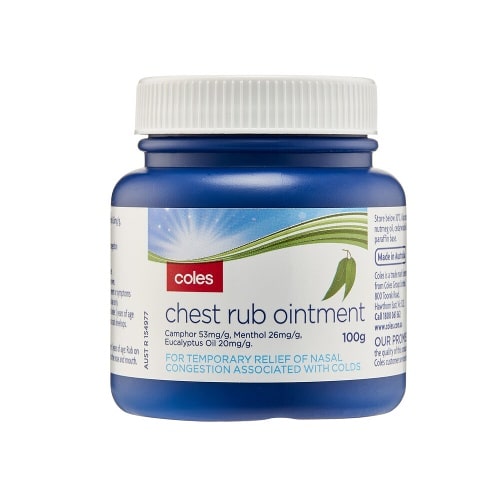}\break{Chest Rub Ointment}&
        \includegraphics[width=0.6\textwidth,trim={4cm 2cm 4cm 2cm},clip]{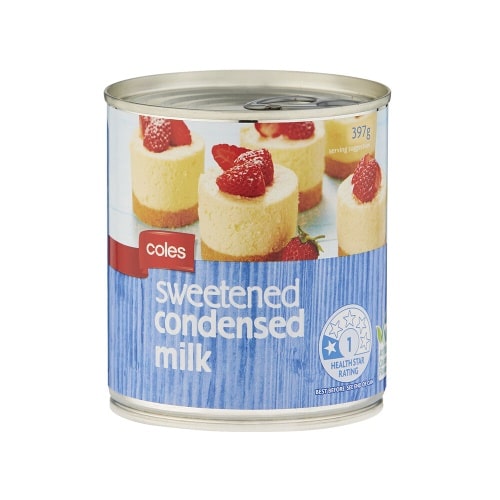}\break{Condensed Milk}&
        \includegraphics[width=0.6\textwidth,trim={4cm 2cm 4cm 2cm},clip]{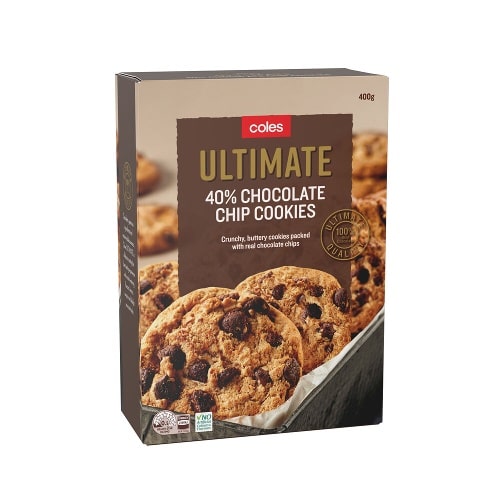}\break{Cookies}&
        \includegraphics[width=0.6\textwidth,trim={2cm 2cm 4cm 2cm},clip]{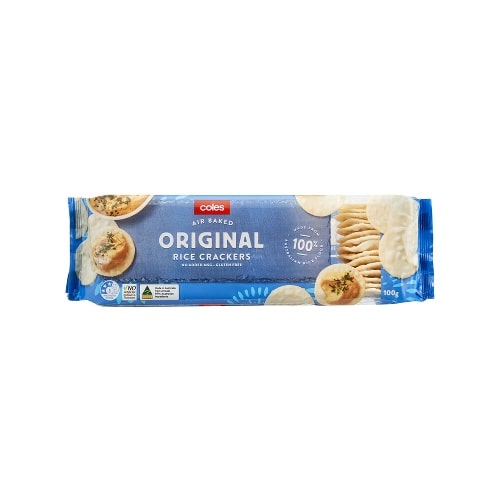}\break{Crackers}&
        \includegraphics[width=0.6\textwidth,trim={3cm 2cm 3cm 2cm},clip]{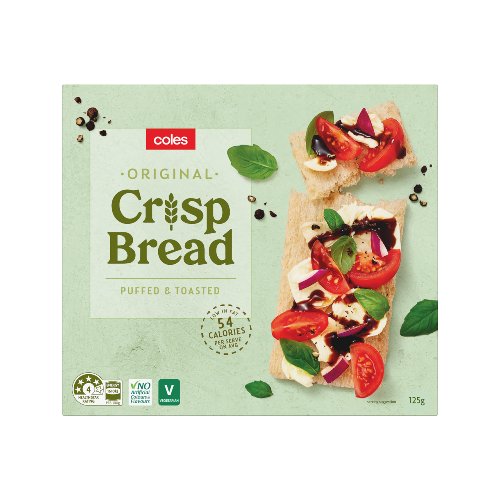}\break{Crisp Bread}\\
        \includegraphics[width=0.6\textwidth,trim={4cm 2cm 4cm 2cm},clip]{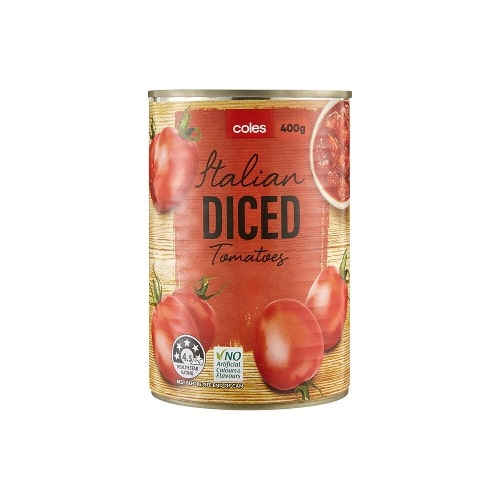}\break{Diced Tomatoes Can}&
        \includegraphics[width=0.6\textwidth,trim={3cm 2cm 3cm 2cm},clip]{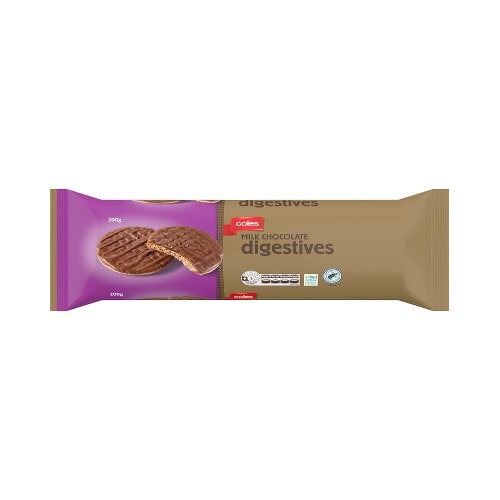}\break{Digestive Biscuits}&
        \includegraphics[width=0.6\textwidth,trim={4cm 2cm 4cm 1cm},clip]{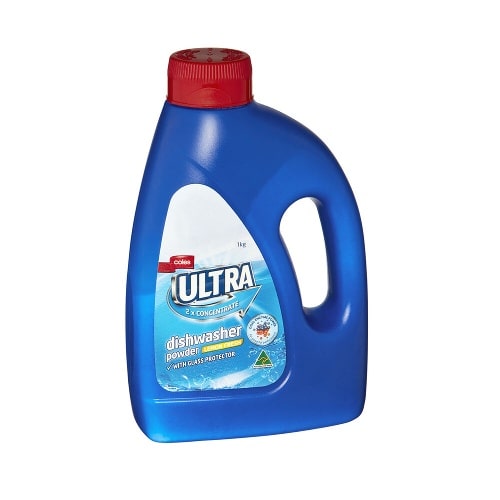}\break{Dishwasher Powder}&
        \includegraphics[width=0.6\textwidth,trim={4cm 2cm 4cm 2cm},clip]{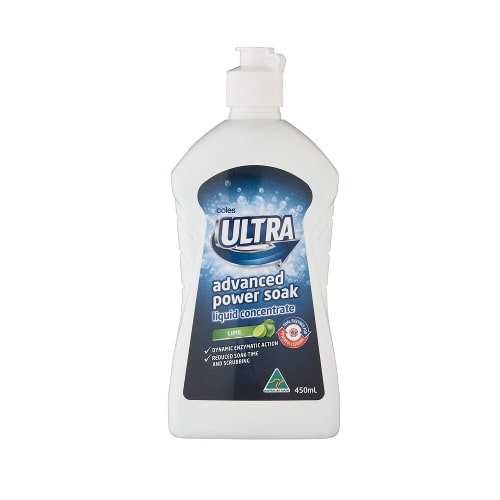}\break{Dishwashing Liquid}&
        \includegraphics[width=0.6\textwidth,trim={4cm 2cm 4cm 2cm},clip]{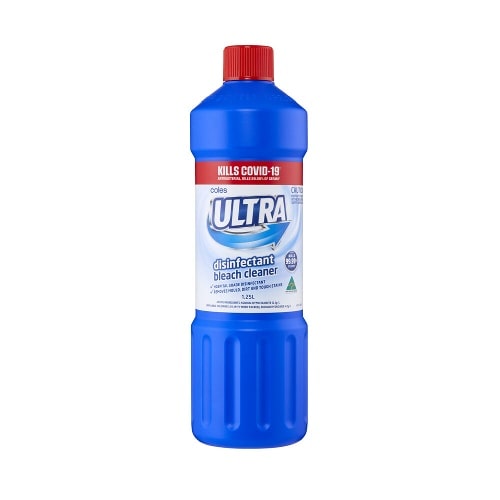}\break{Disinfectant Bleach Cleaner}&
        \includegraphics[trim={70 0 70 0}, clip, width=0.6\textwidth]{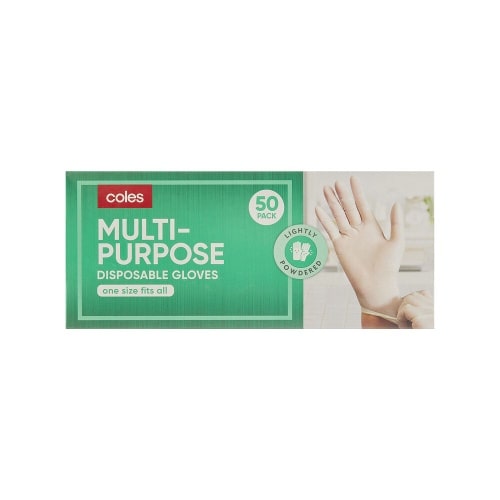}\break{Disposable Gloves}\\
        \includegraphics[width=0.6\textwidth,trim={4cm 2cm 4cm 2cm},clip]{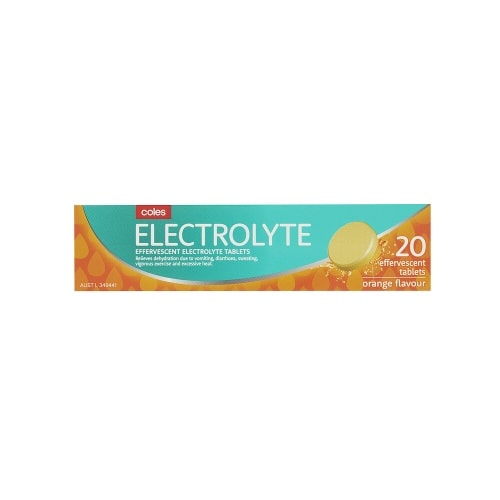}\break{Electrolyte Tablets}&
        \includegraphics[width=0.6\textwidth,trim={4cm 2cm 4cm 2cm},clip]{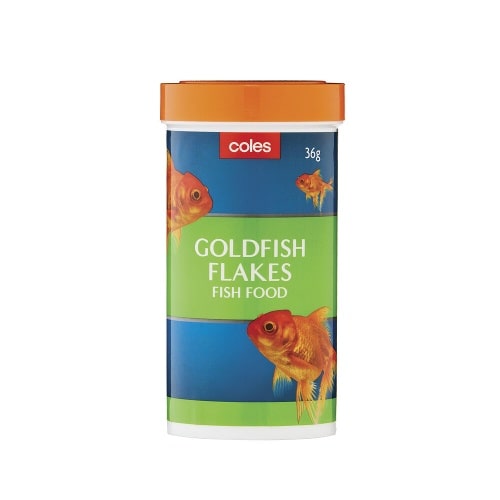}\break{Fish Flakes}&
        \includegraphics[width=0.6\textwidth,trim={4cm 2cm 4cm 2cm},clip]{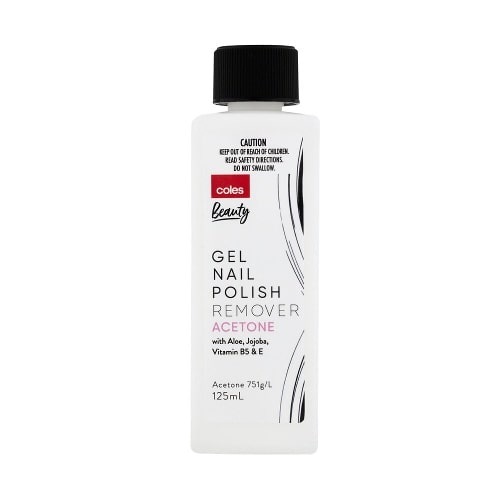}\break{Gel Nail Polish Remover}&
        \includegraphics[width=0.6\textwidth,trim={4cm 2cm 4cm 2cm},clip]{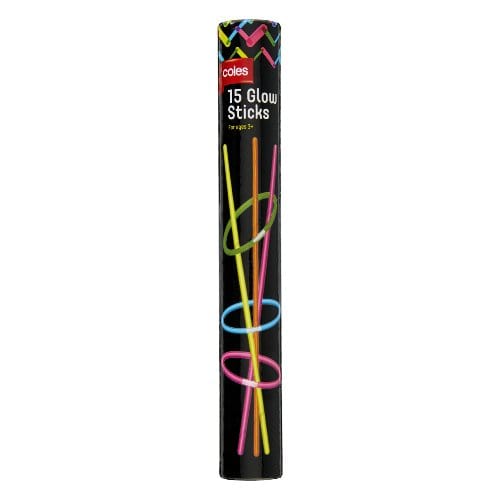}\break{Glow Sticks}&
        \includegraphics[width=0.6\textwidth,trim={4cm 2cm 4cm 1cm},clip]{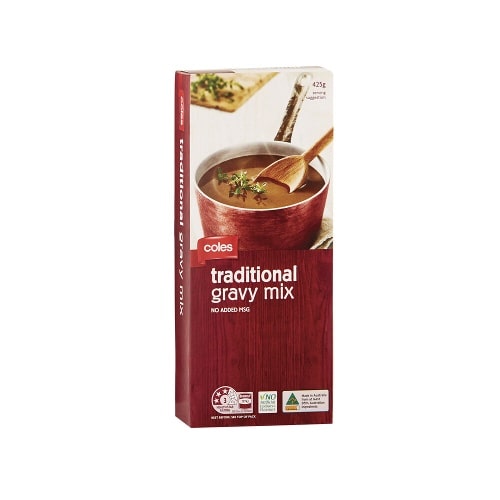}\break{Gravy Mix}&
        \includegraphics[width=0.6\textwidth,trim={4cm 2cm 4cm 2cm},clip]{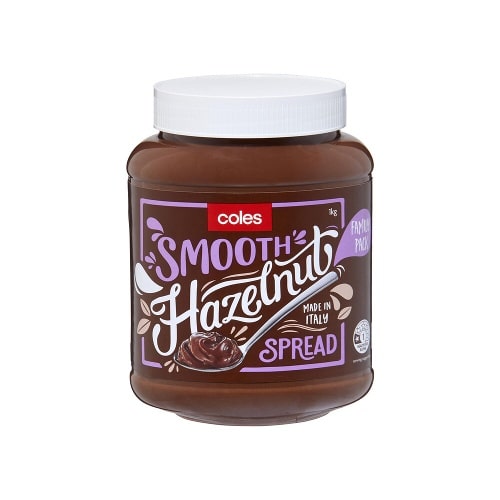}\break{Hazelnut Spread}\\
        \includegraphics[width=0.6\textwidth,trim={2cm 2cm 2cm 2cm},clip]{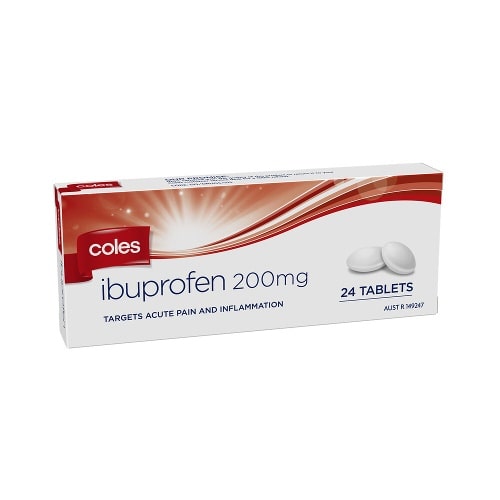}\break{Ibuprofen}&
        \includegraphics[width=0.6\textwidth,trim={4cm 2cm 4cm 2cm},clip]{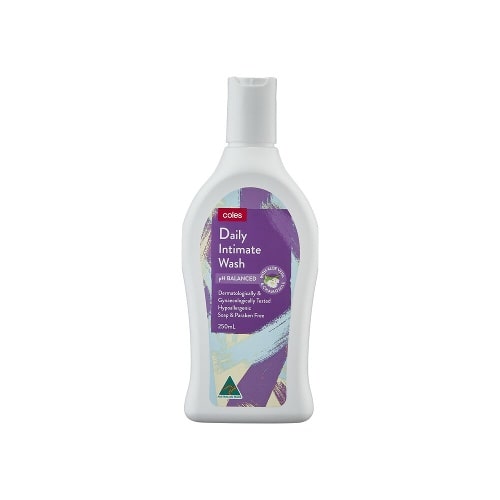}\break{Intimate Wash}&
        \includegraphics[width=0.6\textwidth,trim={4cm 2cm 4cm 2cm},clip]{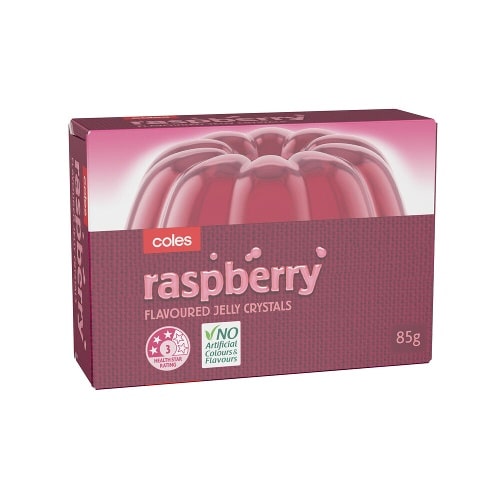}\break{Jelly}&
        \includegraphics[width=0.6\textwidth,trim={4cm 2cm 4cm 2cm},clip]{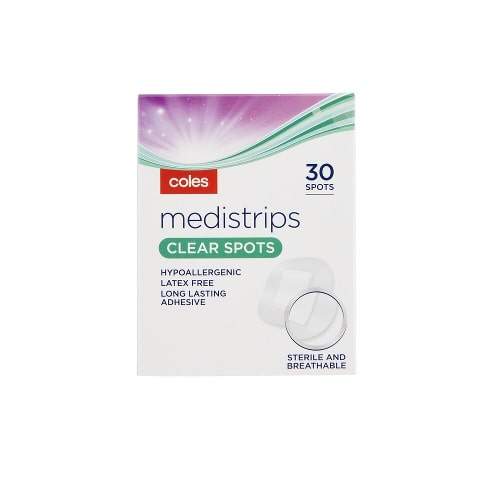}\break{Medistrips}&
        \includegraphics[width=0.6\textwidth,trim={4cm 2cm 4cm 2cm},clip]{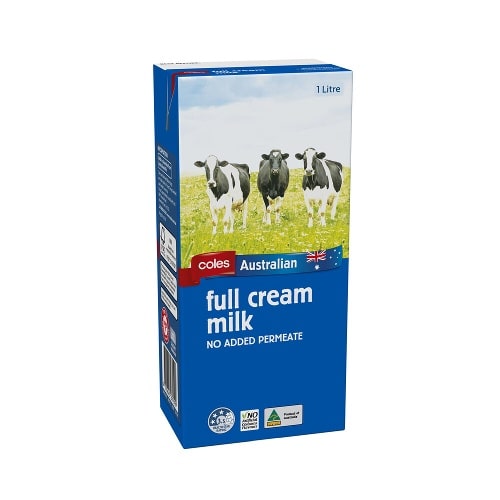}\break{Milk}&
        \includegraphics[width=0.6\textwidth,trim={2.5cm 2cm 3cm 2cm},clip]{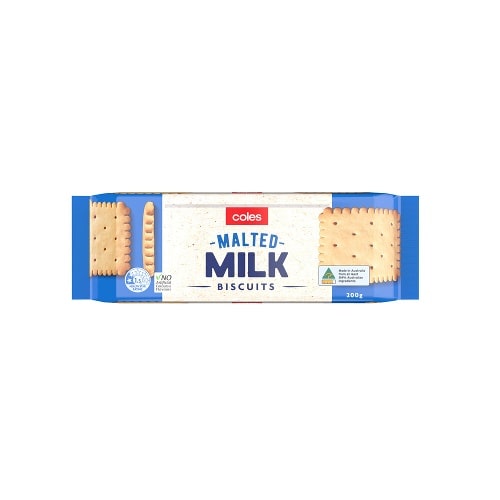}\break{Milk Biscuits}\\
        \includegraphics[width=0.5\textwidth,trim={4cm 0 4cm 2cm},clip]{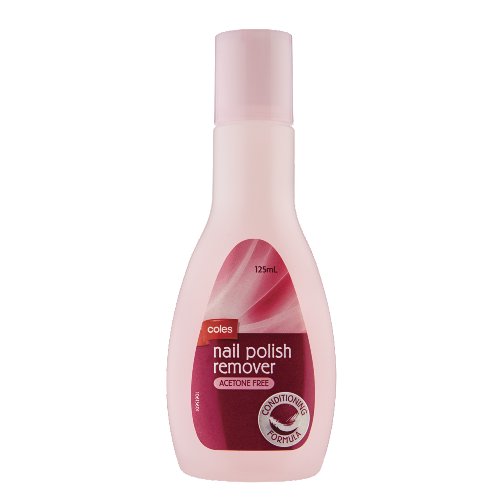}\break{Nail Polish Remover}&
        \includegraphics[width=0.6\textwidth,trim={4cm 2cm 4cm 2cm},clip]{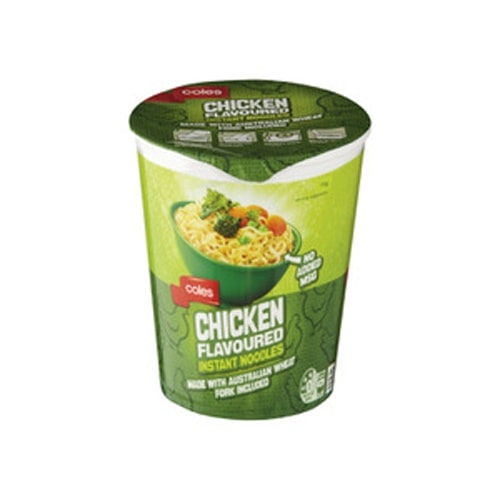}\break{Noodle Cup}&
        \includegraphics[width=0.6\textwidth,trim={3cm 2cm 3cm 2cm},clip]{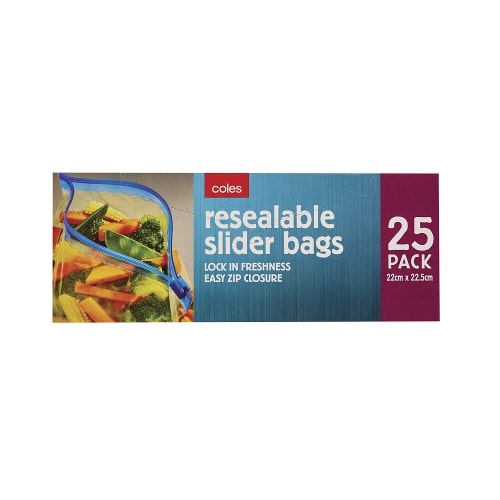}\break{Resealable Bags}&
        \includegraphics[width=0.6\textwidth,trim={2cm 2cm 4cm 2cm},clip]{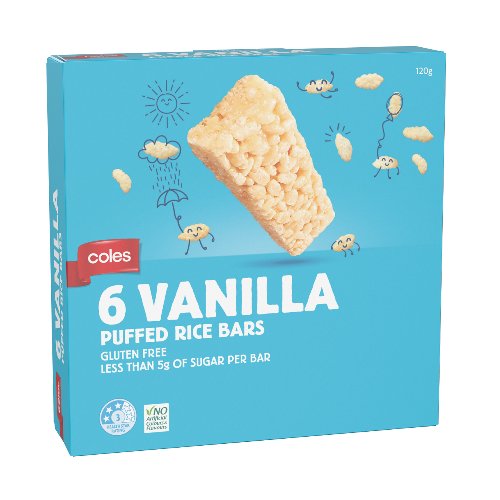}\break{Rice Bars}&
        \includegraphics[width=0.6\textwidth,trim={4cm 1cm 4cm 2cm},clip]{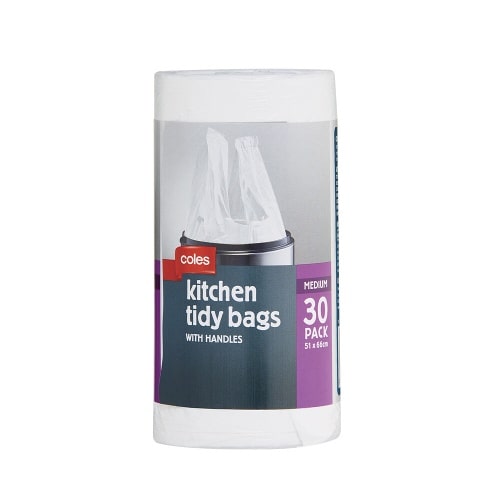}\break{Rubbish Bags}&
        \includegraphics[width=0.6\textwidth,trim={4cm 2cm 4cm 2cm},clip]{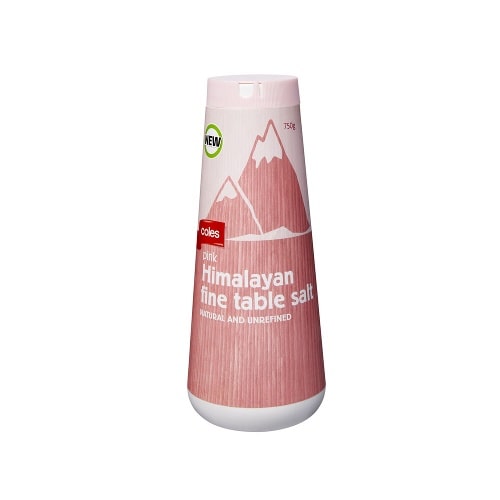}\break{Salt}\\
        \includegraphics[width=0.6\textwidth,trim={4cm 2cm 4cm 2cm},clip]{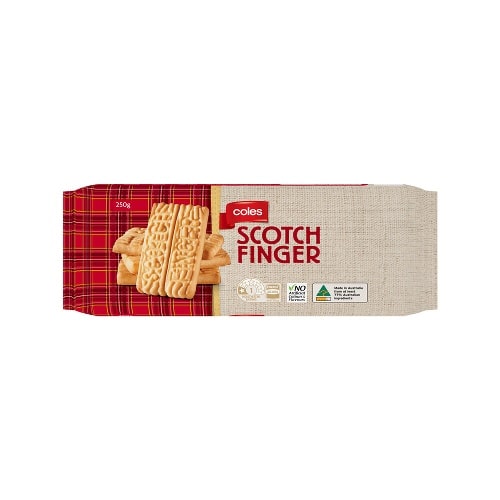}\break{Scotch Finger Biscuits}&
        \includegraphics[width=0.6\textwidth,trim={3cm 2cm 3cm 2cm},clip]{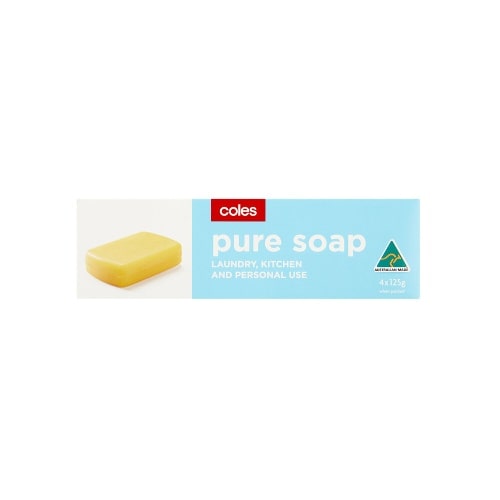}\break{Soap Box}&
        \includegraphics[width=0.6\textwidth,trim={4cm 2cm 4cm 2cm},clip]{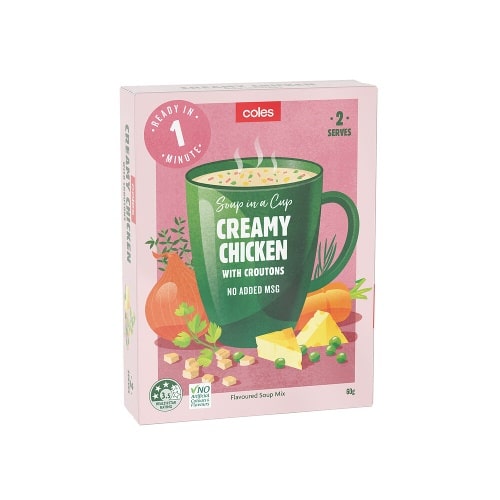}\break{Soup Box}&
        \includegraphics[width=0.5\textwidth,trim={4cm 0cm 4cm 0cm},clip]{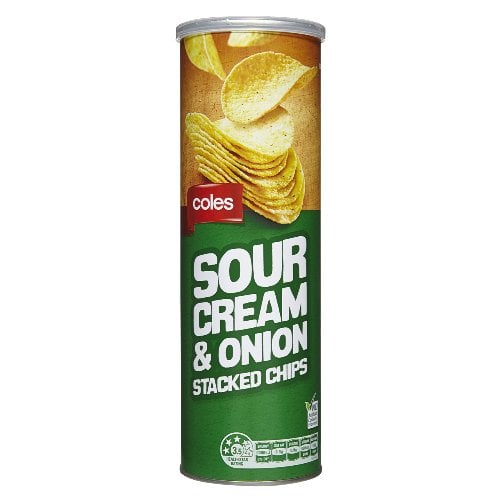}\break{Stacked Chips}&
        \includegraphics[width=0.5\textwidth,trim={4cm 0cm 4cm 0cm},clip]{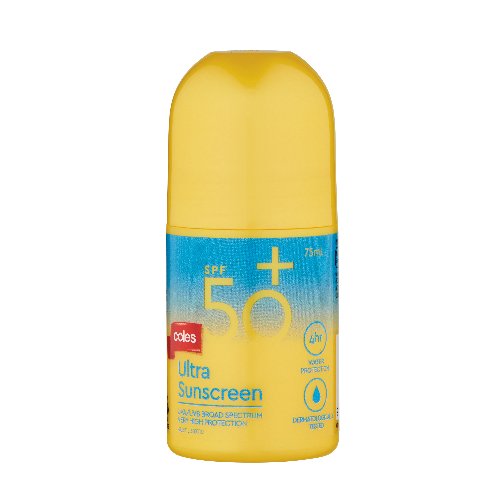}\break{Sunscreen Roll On}&
        \includegraphics[width=0.5\textwidth,trim={4cm 1cm 4cm 2cm},clip]{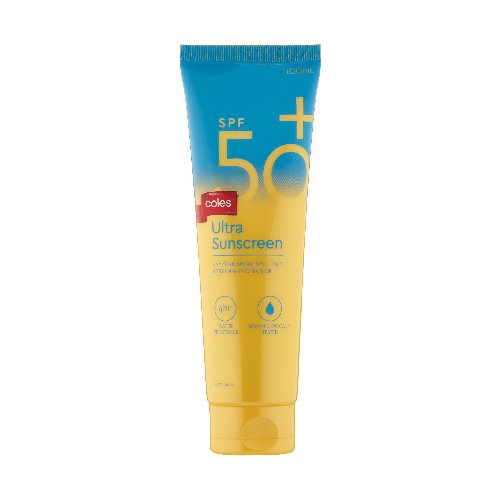}\break{Sunscreen Tube}\\
        \includegraphics[width=0.6\textwidth,trim={4cm 2cm 4cm 2cm},clip]{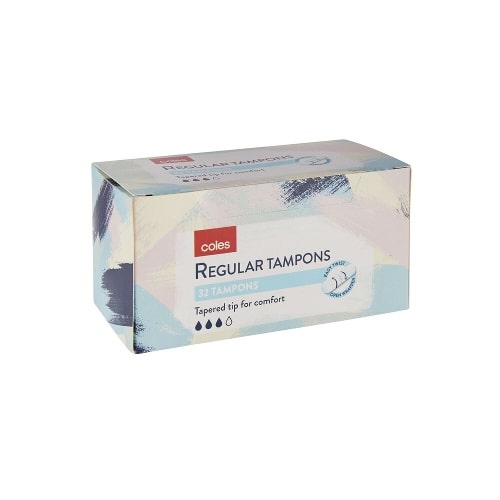}\break{Tampons}&
        \includegraphics[width=0.5\textwidth,trim={4cm 2cm 4cm 2cm},clip]{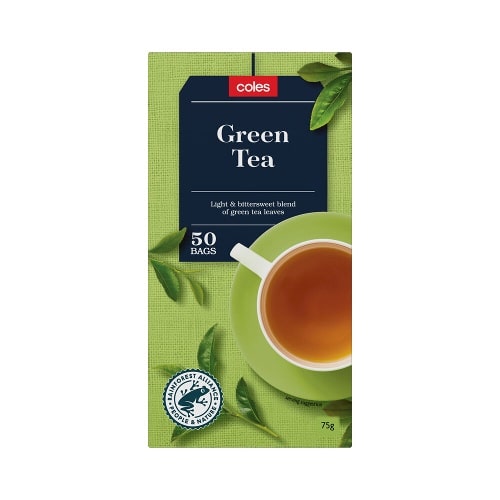}\break{Tea}&
        \includegraphics[width=0.5\textwidth,trim={4cm 1cm 4cm 2cm},clip]{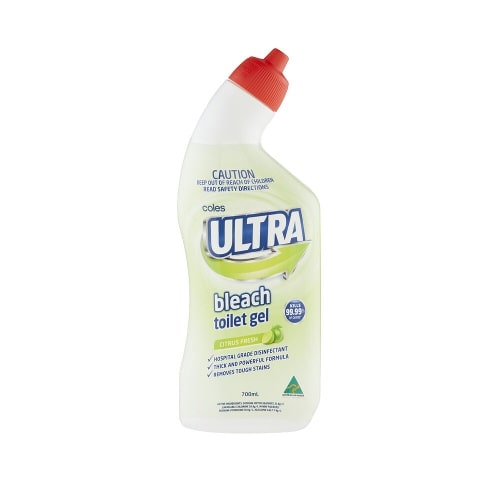}\break{Toilet Cleaner}&
        \includegraphics[width=0.5\textwidth,trim={4cm 2cm 4cm 2cm},clip]{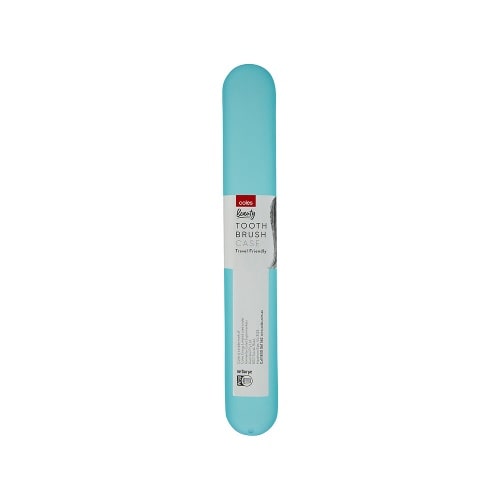}\break{Toothbrush Case}&
        \includegraphics[width=0.6\textwidth,trim={2cm 2cm 2cm 2cm},clip]{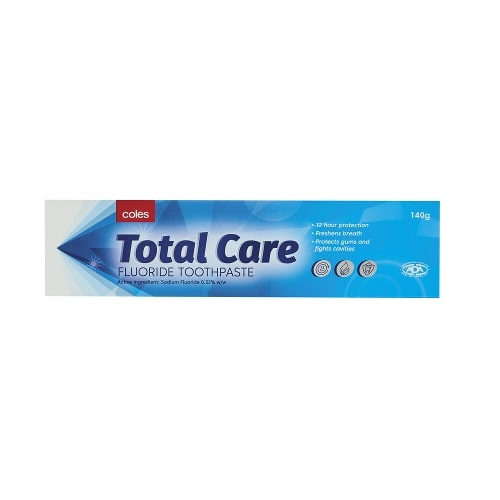}\break{Toothpaste}&
        \includegraphics[width=0.6\textwidth,trim={2cm 2cm 2cm 2cm},clip]{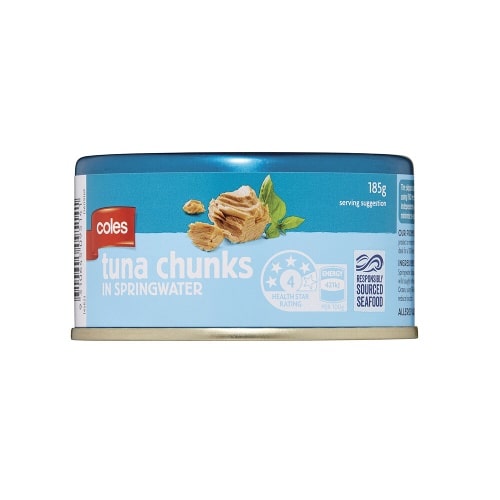}\break{Tuna Can} \\ &&
        \includegraphics[width=0.6\textwidth,trim={2cm 2cm 2cm 2cm},clip]{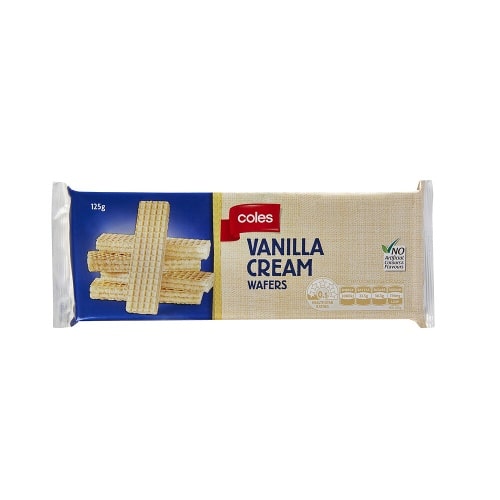}\break{Wafers} &
        \includegraphics[width=0.6\textwidth,trim={2cm 2cm 2cm 2cm},clip]{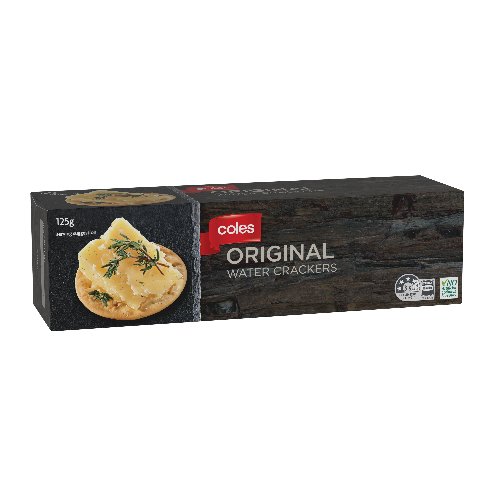}\break{Water Crackers}&&
    \end{NiceTabularX}
    \caption{The 50 objects that comprise the Supermarket Object Set.}
    \label{fig:object_grid}
\end{figure*}

The Supermarket Object Set comprises 50 household items categorized into 10 different categories, with each category containing between 4 and 6 objects. The properties of these objects, including shape, size, and weight, can be found in Table \ref{tab:object_properties}, and the objects are depicted in Figure \ref{fig:object_grid}. The selection of objects for the Supermarket Object Set was guided by several criteria:

\subsection{Cost}
The chosen objects in the Supermarket Object Set are easily obtainable on a budget. This is achieved in three ways: Firstly, all objects can be obtained from Coles, a major Australian supermarket chain, making in-person acquisition simple for local researchers. Secondly, the objects are selected as cheap, generically branded items. Lastly, they are non-perishable and robust, reducing the need for frequent replacement. As a result, the Supermarket Object Set is easily accessible, cost-effective, and durable.

\subsection{Commonality}
The objects in the Supermarket Object Set are chosen from the most commonly purchased generic items at Coles, making the object set representative of standard household objects in Australia. Algorithms that perform effectively on this object set are more likely to exhibit better generalization to real-world Australian settings compared to randomly selected object sets.

\subsection{Shape, Size and Weight}
The graspability of objects is influenced by their size, shape, and weight \cite{morrison2020egad}. To ensure a well-balanced dataset, the objects in the Supermarket Object Set are chosen to cover a diverse range of shapes and sizes. The object categories are explicitly defined based on these properties. For most common household objects, shapes can be categorized as either boxes or cylinders. Therefore, the object set includes separate categories for regular cylinders (e.g., cans), irregular cylinders (e.g., sauce bottles), and boxes (e.g., toothpaste boxes). To ensure size diversity among the selected boxes, separate categories for small and large boxes are included. This distinction is not necessary for cylinders as it was found that the selection criteria for the boxes already provided a wide range of sizes. The weight of an object is also a crucial factor in determining its graspability. Objects in the Supermarket Object Set are carefully selected to represent the range of weights typically encountered in a domestic setting. The dataset includes objects with a diverse set of weights, ranging from light objects weighing a minimum of 18g to heavy objects weighing a maximum of 1458g. The maximum weight is chosen to remain below the maximum payload for most standard robotic manipulators.

\subsection{Variety}
The objects in the Supermarket Object Set are chosen to be representative of a range of household items used in various tasks. This includes non-perishable food items, drinks, cleaning goods, personal hygiene products, and health items. The dataset intentionally includes items with outlier shapes or properties, such as deformable biscuit packets or irregularly shaped spray bottles, to enhance its real-world applicability. The objects are categorized into 10 categories based on their shape, including boxes, cylinders, large objects, and packets.

\subsection{Data Collection Methodology}
\label{sec:data_collection}

\begin{figure}[!ht]
    \centering
    \includegraphics[width=0.4\textwidth]{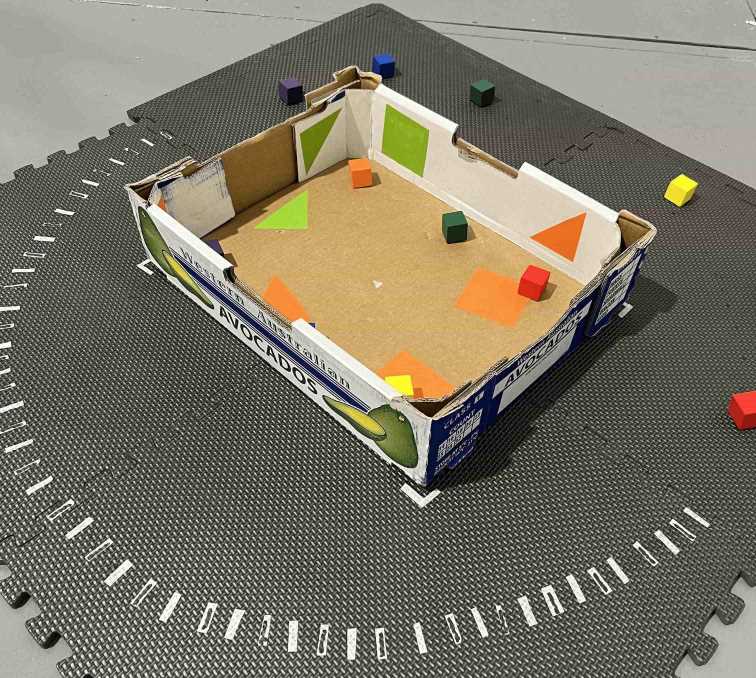}
    \caption{Data collection rig used to image objects in the supermarket object set. The scene is rich with geometries and colours to simplify feature matching.}
    \label{fig:object_ds_rig}
\end{figure}

\begin{figure}[h!]
    \centering
    \includegraphics[width=0.5\textwidth]{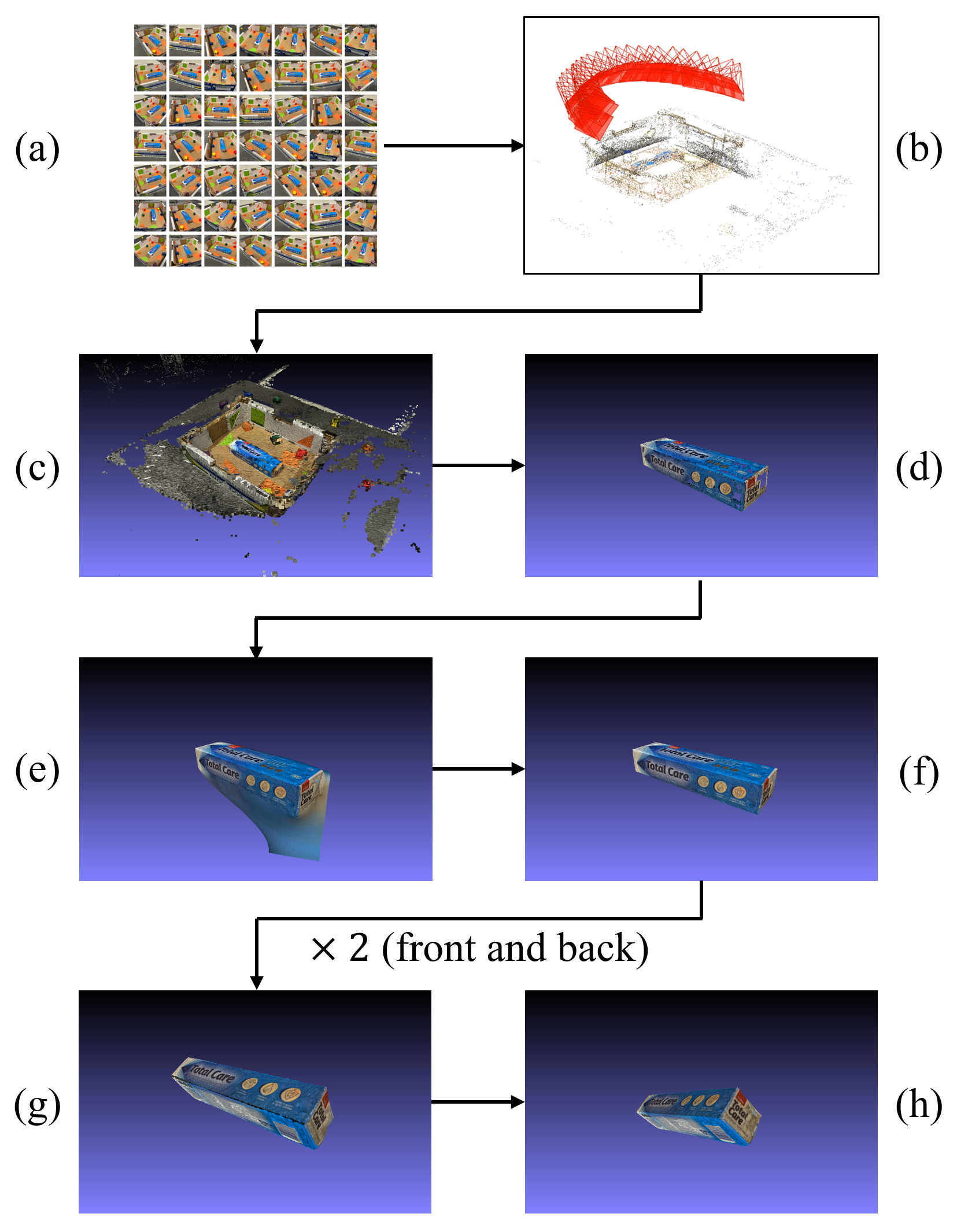}
    \caption{The data collection and post-processing pipeline. (a) Images of the object, (b) camera views (red) and sparse point cloud, (c) dense point cloud, (d) isolated object point cloud, (e) Poisson surface reconstructed mesh, (f) cleaned point cloud, (g) ICP matched mesh halves, (h) final watertight object mesh.}
    \label{fig:object_ds_preprocess}
\end{figure}

To construct the object meshes, a Structure-from-Motion approach called COLMAP \cite{schoenberger2016sfm}, \cite{schoenberger2016mvs} was utilized. The mesh construction process for each object involved the following steps. A diagram illustrating each step of the data collection and post-processing procedure can be found in Figure \ref{fig:object_ds_preprocess}. First, the object was placed in a feature-rich box, as depicted in Figure \ref{fig:object_ds_rig}. The box contained various colored shapes, facilitating easy feature detection and matching using SIFT \cite{Lowe:2004:DIF:993451.996342} and RANSAC \cite{10.1145/358669.358692}. Next, the front and side of the object were imaged using an iPhone 13 mini from 25 different views, covering a semicircle around the object. For each view, a photo of the scene was taken, resulting in a total of 50 photos captured for each object (25 views $\times$ 2 sides). This is depicted in Fig.~\ref{fig:object_ds_preprocess}a and \ref{fig:object_ds_preprocess}b. A high-quality point cloud of the scene was then reconstructed using COLMAP (Fig.~\ref{fig:object_ds_preprocess}c). The object was isolated and cleaned using Meshlab tools (Fig.~\ref{fig:object_ds_preprocess}d). Subsequently, screened Poisson surface reconstruction \cite{10.1145/2487228.2487237} was employed to create a watertight mesh (Fig. \ref{fig:object_ds_preprocess}e). This process introduced some artifacts that were subsequently cleaned (Fig.~\ref{fig:object_ds_preprocess}f). As the bottom of the object, which rests on the table, was not visible and therefore not imaged, the resulting mesh was incomplete. To address this issue and fully reconstruct the mesh, the object was flipped to expose the previously unseen faces, and the aforementioned steps were repeated to generate a second mesh. The two meshes were combined using point-based gluing utilising Iterative Closest Point (ICP) algorithm \cite{10.1007/BF01427149} (Fig.~\ref{fig:object_ds_preprocess}g). This process returns a high-quality water-tight mesh of the object from the object set (Fig.~\ref{fig:object_ds_preprocess}h). This process was undertaken for all 50 objects in the object set to collect our dataset.

\subsection{Metadata}
\label{sec:metadata}
The Supermarket Object Set was constructed from a total of 2,500 images, with each object represented by 50 images. Each image had a resolution of 4032$\times$3024. The final object set consists of 50 files in the Polygon File Format (.ply). The resulting object set consists of 50 files in the Polygon File Format (.ply), which can be loaded using popular computer graphics software such as Meshlab or Blender. When uncompressed, the full dataset occupies a storage space of 14.6 GB.

\section{Conclusions}

The Australian Supermarket Object Set provides a unique and practical addition to the landscape of standardized benchmarking datasets. By offering a collection of 50 easily accessible supermarket items along with high-quality 3D textured meshes, it addresses key challenges faced by researchers, such as accessibility, affordability, and real-world applicability. The dataset enables the benchmarking of algorithms in both simulated and real-world environments while addressing practical challenges like deformability, mass distribution, and durability that are often neglected in existing datasets.

The accessibility of the object set through a local supermarket chain ensures its usability for Australian researchers, while its diversity of shapes, sizes, and categories makes it relevant for a wide range of robotics and computer vision applications. The systematic data collection process, employing high-resolution imaging and Structure-from-Motion techniques, ensures the reliability and accuracy of the dataset.

By focusing on common household items encountered in daily life, the supermarket object set not only serves as a valuable resource for academic research but also has the potential to improve the generalizability of algorithms in real-world settings. We anticipate that this dataset will play a significant role in advancing reproducibility and fostering innovation in robotic manipulation, object detection, and related fields. Future work may expand this dataset to include additional categories or explore its application in domain-specific tasks like grocery sorting or packaging automation.

\section{Acknowledgement}

This research was funded by Coles Group, Australia.

\begin{table*}
\setlength{\tabcolsep}{2pt}
    \centering
    \def\arraystretch{1.3}
    \begin{tabular}{|llll|llll|} \hline
        \textbf{Category} & \textbf{Object} & \textbf{Mass(g)} & \textbf{Dims (mm)} & \textbf{Category} & \textbf{Object} & \textbf{Mass (g)} & \textbf{Dims (mm)}\\\hline
        \multirow{4}{*}{\vtop{\hbox{\strut Small}\hbox{\strut boxes}}}&Alcohol wipes&45&66$\times$107$\times$30&\multirow{5}{*}{\vtop{\hbox{\strut Regular}\hbox{\strut cylinders}\hbox{\strut (food)}}}&Baked beans&485&109$\times$75\\
         &Ibuprofen&18&125$\times$49$\times$19&&Condensed milk&445&84$\times$75\\
         &Medistrips&13&74$\times$96$\times$22&&Diced tomatoes&471&109$\times$75\\
         &Tampons&95&107$\times$54$\times$54& &Stacked chips&183&233$\times$72\\\cline{1-4}
        \multirow{4}{*}{\vtop{\hbox{\strut Small}\hbox{\strut boxes}\hbox{\strut (food)}}}& Electrolyte tablets & 94 & 151$\times$38$\times$38 & &Tuna&126&39$\times$68\\\cline{5-8}
         & Gravy mix & 450 & 80$\times$191$\times$43 &\multirow{5}{*}{\vtop{\hbox{\strut Irregular}\hbox{\strut cylinders}}}&Gel nail polish remover&127&40$\times$124$\times$40\\
         & Jelly & 97 &94$\times$66$\times$29&&Nail polish remover&134&58$\times$146$\times$38\\
         & Soup box & 89 &110$\times$149$\times$30& &Sunscreen roll on&105&107$\times$45\\\cline{1-4}
        \multirow{5}{*}{\vtop{\hbox{\strut Large}\hbox{\strut boxes}}}&Aluminium foil&162&312$\times$51$\times$52& &Sunscreen tube&116&178$\times$36\\
         &Disposable gloves&373&220$\times$97$\times$49& &Toothbrush case&24&30$\times$208$\times$20\\\cline{5-8}
         &Resealable bags&196&237$\times$91$\times$46&\multirow{5}{*}{\vtop{\hbox{\strut Irregular}\hbox{\strut cylinders}\hbox{\strut (food)}}}&BBQ sauce&632&225$\times$68\\
         &Soap box&505&184$\times$62$\times$54&&Burger sauce&477&78$\times$156$\times$58\\
         &Toothpaste&169&206$\times$40$\times$52&&Hazelnut spread&441&82$\times$100$\times$65\\\cline{1-4}
        \multirow{6}{*}{\vtop{\hbox{\strut Large}\hbox{\strut boxes}\hbox{\strut (food)}}}&Cookies&474&149$\times$211$\times$68& &Noodle cup&94&106$\times$95\\
        &Crisp bread&158&162$\times$135$\times$66& &Salt&829&205$\times$83\\\cline{5-8}
        &Milk&1074&92$\times$197$\times$58&\multirow{5}{*}{\vtop{\hbox{\strut Large}\hbox{\strut objects}}}&Bathroom cleaner&654&112$\times$258$\times$53\\
         &Rice bars&170&156$\times$156$\times$39&&Dishwasher powder&1128&140$\times$220$\times$65\\
         &Tea&129&159$\times$83$\times$61& &Dishwashing liquid&510&91$\times$209$\times$43\\
         &Water crackers&161&230$\times$60$\times$60& &Intimate wash&289&67$\times$179$\times$40\\\cline{1-4}
         \multirow{6}{*}{\vtop{\hbox{\strut Regular}\hbox{\strut cylinders}}}&Bubbles&121&102$\times$42& &Toilet cleaner&761&91$\times$246$\times$55\\\cline{5-8}
        &Chest rub ointment&123&68$\times$60&\multirow{5}{*}{Packets}&Crackers&128&265$\times$77$\times$48\\
        &Disinfectant bleach&1458&293$\times$85& &Digestive biscuits&214&280$\times$85$\times$38\\
         &Fish flakes&70&115$\times$60& &Milk biscuits&221&175$\times$64$\times$45\\
         &Glowsticks&96&220$\times$30& &Scotch finger biscuits&257&170$\times$75$\times$44\\
         &Rubbish bags&262&131$\times$56& &Wafers&140&217$\times$79$\times$23\\ \hline
    \end{tabular}
    \caption{Properties of all items in the object set. Dimensions are reported as (horizontal width)$\times$(vertical height)$\times$(depth) for boxes and (height)$\times$(diameter) for cylinders.}
    \label{tab:object_properties}
\end{table*}

\bibliographystyle{named}
\bibliography{refs}

\end{document}